\documentclass[10pt,a4paper]{article}

\usepackage{lrec2006}
\usepackage{covington}
\usepackage{mdwlist}
\usepackage{multirow}
\usepackage{qtree}

\title{Massively Increasing TIMEX3 Resources: A Transduction Approach}
\name{Leon Derczynski$^{\ast}$, H\'ector Llorens$^{\dagger}$, Estela Saquete$^{\dagger}$}

\address{$^{\ast}$University of Sheffield \\ S1 4DP, UK \\ \texttt{leon@dcs.shef.ac.uk} \\ \\
         $^{\dagger}$Universidad de Alicante \\ 03690, Spain \\ \texttt{hector,stela@dlsi.ua.es} }

\abstract{
Automatic annotation of temporal expressions is a research challenge of great interest in the field of information extraction.
Gold standard temporally-annotated resources are limited in size, which makes research using them difficult.
Standards have also evolved over the past decade, so not all temporally annotated data is in the same format.
We vastly increase available human-annotated temporal expression resources by converting older format resources to TimeML/TIMEX3.
This task is difficult due to differing annotation methods.
We present a robust conversion tool and a new, large temporal expression resource. Using this, we evaluate our conversion process by using it as training data for an existing TimeML annotation tool, achieving a 0.87 F1 measure -- better than any system in the TempEval-2 timex recognition exercise.
\\ \newline
}

\begin{document}

\maketitleabstract

\section{Introduction}

In this paper, we introduce a tool for unifying temporal annotations produced under different standards and show how it can be improved to cope with wide variations in language. We then apply our enhanced tool to existing annotated corpora to generate a TIMEX3 corpus larger than the sum of all existing TIMEX3 corpora by an order of magnitude and show that this resource is useful for automatic temporal annotation.

Temporal expressions (\textbf{timexes}) are a basic part of time in language. They refer to a period or specific time, or temporally reify recurrences. Durations such as \emph{``two weeks"} typically have a quantifier and a unit. Dates or times such as \emph{``next Thursday"} and \emph{``July 21 2008"} can be anchored to a calendar and have set beginning and end bounds; sets like \emph{``every summer"} indicate a recurrence pattern. 

Dates can be further broken down into deictic and absolute expressions. Absolute temporal expressions can be directly placed on a calendar without further information, whereas deictic temporal expressions need some external (perhaps anaphoric) information to be resolved. For example, \emph{``April 19"} is deictic, because its year depends on the context in which it appears.

After a decade of development, there are two main standards with which to annotate timexes. \textbf{TIMEX2}~\cite{ferro2004tides} is dedicated to timex annotation. TimeML~\cite{pustejovsky2005specification} is a later standard for all aspects of temporal annotation. It defines \textbf{TIMEX3} for timex annotation, and introduces other entities such as events and temporal relations.

Manual creation of fully temporally annotated resources is a complex and intensive task~\cite{setzer2001pilot}.
This has lead to only small corpora being available. The largest two corpora in TimeML, the current standard for temporal annotation, total about 260 newswire documents including just over 2~000 gold standard TIMEX3 annotations. 
Automatic annotation is also a difficult task, compounded by the scarcity of annotated training data.
To this end, some recent work has explored the complex issue of converting TIMEX2 corpora to TIMEX3~\cite{saquete2011automatic}.

The current state of affairs is that we have small TIMEX3 resources, much larger TIMEX2 resources, and a proof-of-concept tool for mapping from TIMEX2 to TIMEX3. Because data sparsity has limited automatic TimeML and TIMEX3 annotation systems, we assume that increasing the volume of TIMEX3 data will help the performance of such systems. We will do this via conversion of multiple TIMEX2 resources. Our research questions are as follows:

\begin{enumerate}
\item What practical issues are there in converting large-scale TIMEX2 resources to TIMEX3?
\item How can we evaluate the success of such a conversion?
\item Does extra training data help automatic timex annotation?
\end{enumerate}

We answer these questions in this paper. In Section~\ref{previous} we introduce the corpora and an existing format conversion tool and in Section~\ref{method} describe how we enhance it to perform its task more accurately. We use the tool to create the largest current TIMEX3 resource, described in Section~\ref{resource}. We then show how this new training data can be used with a state-of-the-art TIMEX3 annotation system to improve automatic annotation performance in Section~\ref{evaluation} and finally conclude in Section~\ref{conclusion}

\section{Background}
\label{previous}

Manual temporal annotation is a complex, tiring and error-prone process~\cite{setzer2001pilot}. The abstract notion of temporality and the requirement to make formal annotations using time have lead to in-depth annotation schemas accompanied by detailed annotation guidelines. This makes the generation of temporally annotated resources expensive.

Temporal expressions generally fall in to one of four categories. These are:

\label{tempexTypes}
\begin{itemize}
\item \textbf{Absolute} --- Where the text explicitly states an unambiguous time. Depending on the granularity of the interval, the text includes enough information to narrow a point or interval directly down to one single occurrence. This is in contrast to a time which, while precise and maybe easy for humans to pin onto a calendar, relies on an external reference. For example, \emph{Thursday October 1st, 2009} would be considered absolute, but \emph{The week after next} would not - the information is not all explicit or held in the same place; this latter expression implies reliance on some external time point.
\item \textbf{Deictic} --- Cases where, given a known time of utterance, one can determine the period being referred to. These time expressions, specify a temporal distance and direction from the utterance time. One might see a magazine bulletin begin with \emph{Two weeks ago, we were still in Saigon.}; this expression relies on an unclear speech time, which one could safely assume was the date the article was written. More common examples include \emph{tomorrow} and \emph{yesterday}, which are both offset from the time of their utterance.
\item \textbf{Anaphoric} --- Anaphoric temporal expressions have three parts -- temporal distance (e.g. 4 days), temporal direction (past or future), and an anchor that the distance and direction are applied from. The anchor, for anaphoric temporal expressions (sometimes also known as \textbf{relative} temporal expressions), is an abstract discourse-level point.
Example phrases include \emph{the next week}, \emph{that evening} or \emph{a few hours later}, none of which can be anchored even when their time of utterance is known. 
\item \textbf{Duration} --- A duration describes an interval bounded by start and end times. These might be implicit (\emph{during next week}), where the reader must use world knowledge\index{world knowledge} to deduce start and end points and their separation distance, or explicit (\emph{From 8pm to 11.20pm this evening}). Durations generally include a time unit as their head token. This type of temporal expression is easily confused with relative expressions; for example, in \emph{``The plane was flying for seven days"}, the timex \emph{``seven days"} acts as a duration, whereas in  \emph{``I will have finished this in seven days"}, the same timex refers to a point seven days after the utterance.
\end{itemize}

The TIDES TIMEX2 standard~\cite{ferro2004tides}, preceded by the STAG timex descriptions~\cite{setzer2001temporal}, formally defines how to determine what constitutes a temporal expression in discourse and further defines an encoding for temporal expressions. A simple TIMEX2 annotation is shown in Example~\ref{ex:timex2}.

\begin{example}
\scriptsize
\texttt{The Yankees had just finished <TIMEX2 val="1998-10-02TEV">a draining evening</TIMEX2> with a 4-0 decision over the Rangers}
\label{ex:timex2}
\end{example}

The TIMEX2 standard is designed to be the sole temporal annotation applied to a document, and it introduces just one annotation element: \texttt{<TIMEX2>}. As a result, complex time-referring expressions made of contiguous words are labelled as a single TIMEX2, perhaps with specific sub-parts annotated as nested (or ``embedded") TIMEX2s. This is shown in Example~\ref{ex:timex2-nested}.

\begin{example}
\scriptsize
\texttt{before <TIMEX2 VAL="1999-W23">the week of <TIMEX2 VAL="1999-06-07">the seventh</TIMEX2> until <TIMEX2 VAL="1999-06-11">the eleventh</TIMEX2> </TIMEX2>}
\label{ex:timex2-nested}
\end{example}

Later, TIMEX3 was introduced as the next iteration of this timex annotation scheme. As part of TimeML~\cite{pustejovsky2005specification}, which is a rich annotation schema designed to capture a complete range of temporal information, TIMEX3 focuses on minimally-expressed timexes. This means that entities that would have been a nested or event-based temporal expressions are represented as atomic temporal expressions and separate events, the relations between which are described with TimeML TLINKs. In Example~\ref{ex:timex2}, what would have been a single event-based temporal expression under TIMEX3 is broken down into an event and a timex which are co-ordinated by a temporal signal.

\begin{example}
\scriptsize
\texttt{until <TIMEX3 tid="t31" type="DURATION" value="P90D" temporalFunction="false" functionInDocument="NONE">90 days</TIMEX3> <SIGNAL sid="s16">after</SIGNAL> their <EVENT eid="e32" class="OCCURRENCE" stem="issue">issue</EVENT> date.}
\label{ex:timex3}
\end{example}

TimeML removed nested and conjoined timexes, preferring a finer annotation granularity where timexes are events are separate entities with explicit relations defined between them. The work in this paper centres on applying a transducer to TIMEX2 resources to bring them into a LAF-compliant format~\cite{ide2009standards} (our TIMEX3 annotations are valid ISO-TimeML~\cite{pustejovsky2010iso}). The resulting corpora will further the state-of-the-art in temporal information extraction.

\subsection{Comparable Work}

The two most similar previous papers cover generation of TIMEX3 from TIMEX2 resources, and creation of TIMEX3 resources. In this section, we describe them and how our work differs.

Saquete and Pustejovsky~\shortcite{saquete2011automatic} describe a technique for converting TIMEX2 to TIMEX3 annotations and present the \textbf{T2T3} tool as an implementation of it. As some things annotated as TIMEX2s were no longer considered parts of temporal expressions in TimeML and instead assigned to other functions, T2T3 generates not only TIMEX3s but also any extra TimeML elements. T2T3 is evaluated using TimeBank~\cite{pustejovsky2003timebank} and 50 ACE TERN documents.
 This work was novel, but its practical evaluation limited to the TimeBank corpus and a small selection from the ACE TERN data.
In terms of temporal expressions, there is not much more diversity to be found in TimeBank, which is often used as the sole training and testing resource for temporal information processing systems.
Although it is new data, only a small sample of the ACE data was used for the original evaluation of T2T3. In our work, we greatly increase the volume and variety of text converted from TIMEX2, creating a more robust, enhanced tool that works beyond a demonstration dataset.

TimeBank~\cite{pustejovsky2003timebank} and the AQUAINT TimeML corpus\footnote{\scriptsize See \texttt{http://www.timeml.org/timebank/timebank.html}.} comprise around 250 TimeML annotated documents. These newswire corpora have annotations for temporal entities other than timexes in addition to a total of 2~023 TIMEX3 annotations.
While mature and gold-standard annotated, existing TimeML corpora (TimeBank and the AQUAINT TimeML corpus\footnote{\scriptsize See \texttt{http://www.timeml.org/}.}) are limited in size and scope, and larger resources are required to advance the state of the art. Our contribution is that we introduce new high-quality automatically-generated resources, derived from gold-standard annotations. These comprise large numbers of new timex, event and relation annotations, covering a wider range of forms of expression.

\begin{table}
\begin{center}
\begin{tabular}{| l | c | c | r |}
\hline
\textbf{Resource name} & \textbf{Type} & \textbf{Words} & \textbf{Annotations} \\
\hline
TimeBank v1.2 & TIMEX3 & 68.5K & 1 414\\
AQUAINT & TIMEX3 & 34.1K & 609\\
TempEval-2 test & TIMEX3 & 5.5K & 81 \\
TimenEval & TIMEX3 & 7.9K & 214\\ 
\emph{Total} & & 116K & \emph{3 289} \\
\hline
WikiWars & TIMEX2 & 120K & 2 681\\
ACE 2004 TERN & TIMEX2 & 54.6K & 8 047\\
ACE 2005 & TIMEX2 & 260K & 5 483\\
TIDES dialogue & TIMEX2& 31.6K & 3 541\\
\emph{Total} & & 466K & \emph{19 752} \\
\hline
\end{tabular}
\end{center}
\caption{A summary of publicly-available TIMEX2- and TIMEX3-annotated corpora for English.}
\label{tab:corpora}
\end{table}

\subsection{TIMEX2 Datasets}

There are a few TIMEX2-standard datasets available, both new and old. In this section, we describe the publicly-available TIMEX2-annotated corpora.

Corpora are still produced in TIMEX2 format~\cite{mazur2010wikiwars,stroetgen2011wikiwars}. It is less complex than TimeML and if one is only concerned with temporal expressions, one may annotate these adequately without requiring annotation of temporal signals, events and relations. This gives the situation where similar information is annotated in incompatible formats, impeding the work of those interested in TimeML annotation.

TIMEX2 resources were produced in volume for the ACE TERN tasks~\cite{ferro2004tern} and temporal information extraction research conducted shortly after. These contained no other temporal annotations (e.g. for events). Considerable investment was made in developing annotation guidelines and resources and as a result some very large and well-annotated corpora are available in TIMEX2 format. For example, the \textit{ACE 2004 Development Corpus}\footnote{\scriptsize See LDC catalogue refs. \texttt{LDC2005T07} \& \texttt{LDC2010T18}.} contains almost 900 documents including approximately 8~000 TIMEX2 annotations. For a discussion of the nuances of these resources and this standard, see Mazur and Dale~\shortcite{mazur2010wikiwars}.

The ACE2005 corpus\footnote{\scriptsize See LDC catalogue ref. \texttt{LDC2006T06}.}~\cite{strassel2008linguistic} includes text of multiple genres annotated with a variety of entity types, including timexes. The corpus contains text from broadcast news, newswire, web logs, broadcast conversation, usenet discussions, and conversational telephone speech -- a much wider range of genres than existing English TIMEX3 resources (which are almost exclusively newswire).

As part of an effort to diversify the genres of timex-annotated corpora, WikiWars~\cite{mazur2010wikiwars} is a 20-document corpus of Wikipedia articles about significant wars, annotated to TIMEX2. Document length provides interesting challenges regarding tracking frame of temporal reference and co-reference, and the historical genre provides a wide range of temporal granularities (from seconds to millenia) as well as a wealth of non-contemporary timexes.

Finally, the TIDES Parallel Temporal corpus contains transcriptions of conversations about arranging dates. The conversations were originally in Spanish and comprised that language's part of the Enthusiast corpus~\cite{Suhm94speech}, which were later translated into English (by humans). These dialogues thus comprise a parallel corpus rich in temporal language, where both languages are fully annotated according to TIMEX2. Utterances in this case tend to have a high ratio of timexes per sentence, and the language used to describe times is heavily context-dependent compared to newswire. For example, dates and times are often referred to by only numbers (\emph{``How about the ninth? Or the tenth?"} without an accompanying explicit temporal unit.

A summary of timex-annotated English corpora is given in Table~\ref{tab:corpora}. Aside from TimeBank and AQUAINT, other relevant TIMEX3 corpora are the TempEval-2 international evaluation exercise dataset~\cite{verhagen2010semeval} and the TimenEval TIMEX3 dataset~\cite{llorens2012timen}. 

\begin{figure*}
\begin{example}
\scriptsize
\begin{verbatim}
<timex2 ID="TTRACY_20050223.1049-T1" VAL="FUTURE_REF" ANCHOR_VAL="2005-02-23T10:49:00" ANCHOR_DIR="AFTER">
  <timex2_mention ID="TTRACY_20050223.1049-T1-1">
    <extent>
      <charseq START="1768" END="1787">the next month or so</charseq>
    </extent>
  </timex2_mention>
</timex2>
\end{verbatim}
\label{ex:ace}
\end{example}
\caption{Example ACE2005 corpus standoff annotation.}
\end{figure*}

\subsection{Applications}

Here we discuss three applications of the resulting TIMEX3 resource: improved timex recognition, improved timex interpretation and temporal annotation of the semantic web.

Annotating non-newswire texts is problematic with only newswire training data, and solving this problem has practical benefits. TIMEX3 annotated resources are almost exclusively newswire, and the breadth of genres covered by TIMEX2 resources should help with this problem. These previous datasets cover a wide variety of genres, as opposed to existing TIMEX3 resources, which are (with the partial exception of three TimenEval documents) all newswire. The limited variation in forms of expression given a single genre reduces performance of timex recognition systems trained on such data when applied to other genres. Thus, our addition of TIMEX3 annotations in new genres should permit improvements in timex annotation performance in more general contexts.

The ability to automatically build a formal representation of a temporal expression from a phrase in text is improved with more source data. After a timex's extents have been determined, the next annotation step is to interpret it in context and build a standardised representation of timex's semantics, such as an ISO 8601 compliant specification of a calendrical time or date. This is called timex normalisation. In the small existing datasets, newswire, dates, times and durations are expressed in a limited manner. The diversity of temporal expression phrases grows with the volume of annotated timex resources. Building a complete and high-performance temporal expression normalisation system therefore requires a large and diverse resource.

The Semantic web poses a tough temporal annotation problem~\cite{wilks2008semantic}. To temporally annotate the semantic web, one requires both a standard and also tools capable of performing reliable annotation on data with extremely variable quality. Annotation standards have been proposed -- TIMEX3 is suitable for temporal expressions, and OWL-TIME~\cite{hobbs2004ontology} is a temporal ontology suitable for the semantic web. When it comes to dealing with text quality on the web, even semi-structured resources such as Wikipedia pose challenges~\cite{volkel2006semantic,maynard2009sprat,wang2010timely}. For example, dates are often expressed inconsistently on Wikipedia as well as other phrases used to express durations, times and sets, both in article text and infoboxes. While a capable timex normalisation system should be able to handle variances in this kind of expression, the lack of formal timex annotation can make for slow work. Thanks to WikiWars, our final TIMEX3 resource includes a significant amount of Wikipedia data, annotated and normalised in TIMEX3 format. This paves the way for the creation of data-driven systems that are capable of formally annotating Wikipedia (and other resources) for the semantic web.

\section{Method}
\label{method}

The original T2T3 tool worked well with a subset of the ACE TERN corpus and TimeBank. However, upgrades were needed to cope with linguistic variations in new text. In this section, we detail our handling of the source datasets and our solutions to linguistic and technical shortcomings of the original T2T3 when applied to these datasets.

Our general approach has three stages. Firstly, we pre-process the source documents into a uniform format. Then, we run T2T3 over each document individually. Finally, we wrap the resulting annotations in TimeML header and footer and validate them. This process produces a corpus based on gold-standard annotations, though cannot be said to be gold-standard as the machine-generated annotation transductions have not all been manually checked and corrected. To compensate for this, we release the corpus as version 1.0, and will provide future releases repairing mis-annotations as they are found.

Our development cycle consisted of processing source documents with T2T3 and then validating the output using a TimeML corpus analysis tool~\cite{derczynski2010analysing}. We would then compare the structure of the source documents with the consequent TimeML. Any errors or mis-conversions prompted modifications to T2T3. Converting WikiWars proved an especially useful challenge due to the variety of non-English text and encodings found within.

In this section we describe our TIMEX2 corpus preprocessing, the enhancements made to T2T3, and the validation process.

\subsubsection{Preprocessing}

The target format for T2T3 to work with is plain Unicode text, containing TIMEX2 annotations delimited by \texttt{<TIMEX2>} tags. The following work needed to be done to bring source corpora into this format. All meta-information and other XML tags are stripped. In the case of the ACE2005 data, standoff annotations such as in Example~\ref{ex:ace}. Along with the source documents, these annotations were merged in to form inline TIMEX2 elements. Finally, all documents were (where possible) converted to UTF8 or UTF16, with unrecognised entities removed. Wikiwars documents were the hardest to map, having more than one encoding, but these contain words from almost twenty languages in total with more than seven different writing systems.

\begin{figure}
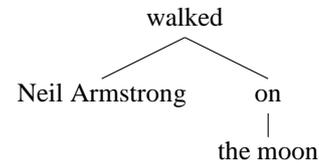

%\centering
\Tree [.walked [.{Neil Armstrong} ] [.on [.{the moon} ] ] ]
\caption{A chunk of a sentence, dependency parsed in order to find which word to annotate as an event.}
\label{fig:event-based-parse}
\end{figure}

\subsection{Running T2T3}

\subsubsection{Signalled event-based times}

Some longer TIMEX2s position a timex relative to an event by means of a co-ordinating phrase with temporal meaning. This co-ordinating phrase is known as a temporal \textbf{signal}. To convert this into TimeML, the event and signal need to be identified, allowing shortening of annotation to just the timex according to the standard. For this, we use an approach that first identifies the signal (according to  the definition and investigation of temporal signals provided in  \newcite{derczynski2011sigcorp}) and then determines which parts of the remaining parts of the phrase (``chunks") are a TimeML TIMEX3 and EVENT.

This procedure constitutes handling a special case (also the majority case) of event-based times, where an event provides a deictic reference required to normalise the time expression. Example~\ref{ex:event-based} is a single TIMEX2, whereas the only TIMEX3 in the phrase is \emph{Tuesday}.

\begin{example}
\emph{``The Tuesday after the party"}
\label{ex:event-based}
\end{example}

The example might look like this as a TIMEX2:

\begin{example}
\scriptsize
\texttt{
<TIMEX2 VAL="2012-03-20">The Tuesday after the party</TIMEX2>
}
\end{example}

and as follows in (slightly simplified) TimeML:

\begin{example}
\scriptsize
\begin{verbatim}
The
<TIMEX3 tid="t1" value="2012-03-20">Tuesday</TIMEX3>
<SIGNAL sid="s1">after</SIGNAL>
the
<EVENT eid="e1" type="OCCURRENCE">party</EVENT>
<TLINK timeID="t1" relType="AFTER"
 relatedEventID="e1" signalID="s1" />
\end{verbatim}
\label{ex:signal-timex3}
\end{example}

Example~\ref{ex:signal-timex3} shows the expansion of a signalled event-based TIMEX2 into TimeML EVENT, SIGNAL, TLINK and TIMEX3 annotations. One may unpack Example~\ref{ex:event-based} as follows: \emph{the party} is an event, \emph{Tuesday} a TIMEX3 and \emph{after} a temporal signal that explicitly connects the TIMEX3 and event, using a TimeML TLINK. 

To achieve this kind of unpacking, it is critical to first select the signal correctly and then subdivide the remainder of the TIMEX2 annotation in order to determine the event and timex elements. We approach this as follows.

\begin{enumerate}
\item From a closed class of temporal signal phrases, find a phrase that co-ordinates the TIMEX2. Our strategy in the case that there is more than one candidate is this. Based on a corpus-based survey of temporal signal phrase meanings~\cite{derczynski2011sigcorp}, we prefer monosemous words (giving preference to the most frequently-occurring ones) followed by polysemous words in descending order of likelihood of being a temporal signal. This gives us at most one signal annotation.
\item Split the original timex into (up to) three contiguous chunks: pre-signal words, signal phrase, and post-signal words.
\item Make the timex chunk the shortest one that has a timex measure word (such as \emph{``day"}), removing tailing or prefixing prepositions and articles. If there is no such matching chunk, make the first chunk the timex chunk.
\item The event chunk contains an event word. Annotate the word that dominates this chunk, based on a dependency parse~\cite{de2006generating} -- see Figure~\ref{fig:event-based-parse}.
\item Add an untyped TLINK between the event and timex, supported by the signal.
\end{enumerate}

For example, in \emph{``the 30 years since Neil Armstrong walked on the moon"}, we split on the monosemous signal word \emph{since} (and not \emph{on}). The time chunk is initially \emph{the 30 years}, from which we remove \emph{the} to end up with \emph{30 years} -- the destination TIMEX3, given the same value as in TIMEX2 (a duration, \texttt{P30Y}). The remainder is dependency parsed (Figure~\ref{fig:event-based-parse}) and the dominant word, \emph{walked}, annotated as an event.

\begin{figure*}
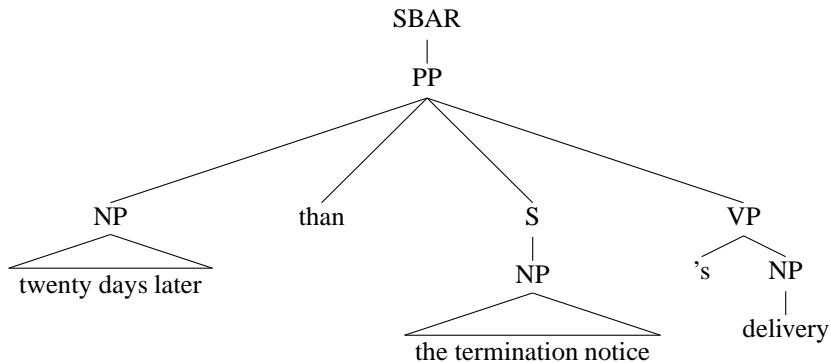

\Tree [.SBAR [.PP \qroof{twenty days later}.NP than [.S \qroof{the termination notice}.NP ] [.VP 's [.NP delivery ] ] ] ]
\caption{Constituent parse of a long TIMEX2.}
\label{fig:long-tree}
\end{figure*}

\begin{table}
\begin{center}
\begin{tabular}{ | r | l | }
\hline
\textbf{Timex length} & \textbf{Frequency} \\
\hline
1 & 654\\
2 & 426\\
3 & 226\\
4 & 81\\
5 & 22\\
6 & 1\\
$\ge$7 & 0\\
\hline
\end{tabular}
\end{center}
\caption{Distribution of token-length of TIMEX3s in TimeBank.}
\label{tab:timex3-lengths}
\end{table}

\subsubsection{Nested expressions}
As discussed in Section~\ref{previous}, TIMEX2 produces larger annotations than 3, which may be nested (as in Example~\ref{ex:timex2-nested}). T2T3 does not handle these. They need to be mapped to multiple TIMEX3 annotations, perhaps with an associated \texttt{anchorTimeID} attribute or temporal relation.

Following from that above example, given the text of \emph{``the week of the seventh"}, the destination TimeML annotation is to describe a week-long duration, two single specific days, and the temporal relations between all three. This would look as follows:

\begin{example}
\scriptsize
\begin{verbatim}
<TIMEX3 tid="t1" type="DATE" value="1999-W23">the
 week</TIMEX3>
<SIGNAL sid="s1">of</SIGNAL>
<TIMEX3 tid="t2" type="DATE" value="1999-06-07">
 the seventh</TIMEX3>

<TLINK timeID="t1" relType="INCLUDED\_BY"
 relatedToTime="t2" signalID="s1" />
\end{verbatim}
\end{example}

We reach this automatically by:

\begin{enumerate}
\item Finding all the TIMEX2s in the scope of the outer one which do not have any children, and mapping them to TIMEX3;
\item Searching for co-ordinating phrases indicating temporal relations, and annotating those as signals;
\item Break the string into chunks, with boundaries based on tokens and sub-element -- new TIMEX3 and SIGNAL -- bounds;
\item Select the chunk most likely to be a timex corresponding to the TIMEX2 \texttt{VAL} attribute, preferring chunks containing temporal measure words (such as \emph{week}) and chunks near the front, and convert it to TIMEX3;
\item Insert TLINK annotations to co-ordinate the new elements, based on value clues and signal-suggested orderings (relation type is left blank when ambiguous).
\end{enumerate}

\subsubsection{Brevity}

We automatically trim long timexes. TIMEX3 annotations are minimal -- that is, including the minimal set of words that can describe a temporal expression -- where TIMEX2 can include whole phrases.  Even after reducing the long annotations that contain temporal substructres, a significant amount of text can remain in some cases. To handle this, we implement reduction of long TIMEX2s into just the TIMEX3-functional part. This is done by measuring the distribution of TIMEX3 token lengths in gold standard corpora, and determining a cut-off point. This distribution is shown in Table~\ref{tab:timex3-lengths}. Any TIMEX2s of six tokens or more that have no yet been handled by the algorithms mentioned above are syntactically parsed. They are then reduced to the largest same-constituent chunk that is shorter than six tokens and contains a temporal measure word, with preference given to the leftmost arguments.

Example~\ref{ex:long} shows a long TIMEX2.

\begin{example}
\emph{twenty days later than the termination notice's delivery}
\label{ex:long}
\end{example}

This produces the constituent tree shown in Figure~\ref{fig:long-tree}. In this case, the four chunks below the root node are considered first; the NP contains a temporal measure word (\emph{days}) and so the TIMEX2 annotation over the whole string is reduced to a TIMEX3 over just \emph{``twenty days later"}.

\subsubsection{Technical changes}
To speed up processing, we moved to NLTK\footnote{\scriptsize See \texttt{http://www.nltk.org/}.} for PoS tagging, which is a maximum entropy based tagger trained on the Penn Treebank. We also stopped doing lemmatisation, as in practise it is never used. For a further speedup, we implemented a timex phrase PoS tagging cache; this reduced execution times by two orders of magnitude.

The tool has generally become more robust and now handles a greater range of texts, providing more precise TimeML annotation. Our work has resulted in a publicly available tool, downloadable from a public Mercurial repository\footnote{\scriptsize See \texttt{http://bitbucket.org/leondz/t2t3}.}.

\subsection{Document post-processing}

After conversion of the body text to TIMEX3, each document is designated at TimeML by giving it an XML header and wrapping the text in \texttt{<TimeML>} elements. Each document is then processed by a DOM parser to check for basic validity, then strictly evaluated compared to the TimeML XSD to check for representation errors, and finally verified at a high level with CAVaT~\cite{derczynski2010analysing}. This results in consistent and TimeML-valid documents.

\begin{table}
\begin{center}
\begin{tabular}{ | l | r | r | }
\hline
\textbf{Corpus name} & \textbf{TIMEX2s} & \textbf{TIMEX3s} \\
\hline
WikiWars & 2 681 & 2 676 \\
ACE 2004 TERN & 8 047 & 7 638  \\
ACE 2005 & 5 483 & 5 199 \\
TIDES parallel dialogue & 3 481 & 3 290 \\
\hline
\end{tabular}
\end{center}
\caption{Annotation counts after conversion to TIMEX3.}
\label{tab:result-timeml}
\end{table}

\begin{table}
\begin{center}
\begin{tabular}{| l | r | r | r | r |}
\hline
\textbf{Corpus} & \textbf{DATE} & \textbf{DUR.} & \textbf{TIME} & \textbf{SET} \\
\hline
TimeBank & 1 164 & 175 & 63 & 12 \\
AQUAINT & 495 & 69 & 27 & 14 \\ 
TempEval-2 & 984 & 170 & 42 & 12 \\
TimenEval & 121 & 34 & 43 & 16 \\
\hline
WikiWars & 2 323 & 230 & 93 & 30  \\
ACE 2004 & 4 697 & 947 & 1 759 & 235 \\
ACE 2005 & 3 024 & 628 & 1 406 & 141 \\
TIDES dialogue & 1 684 & 141 & 1 402 & 63\\
\hline
\emph{Total} & \emph{14 492} & \emph{2 394} & \emph{4 835} & \emph{523} \\
\hline
\end{tabular}
\end{center}
\caption{Distribution of timex types in all TIMEX3 corpora.}
\label{tab:result-types}
\end{table}

\begin{table*}
\centering
\begin{tabular}{|l|l|c|c|c|c|c|c|}
\hline
\multicolumn{2}{|c|}{\textbf{Corpus}} & \multicolumn{3}{|c|}{\textbf{TempEval-2}} & \multicolumn{3}{|c|}{\textbf{Entity-based}} \\
\hline
\textbf{Train} & \textbf{Test} & \textbf{P} & \textbf{R} & \textbf{F1} & \textbf{P} & \textbf{R} & \textbf{F1} \\
\hline
TBAQ & T2T3 & 0.84 & 0.35 & 0.49 & 64.4\% & 32.9\% & 43.6\% \\
TBAQ + T2T3 & 80/20 split & 0.84 & 0.74 & 0.79 & 72.1\% & 71.0\% & 71.5\% \\
TBAQ train (80\%) & TBAQ test (20\%) & 0.93 & 0.80 & 0.86 & 85.2\% & 69.8\% & 76.7\% \\
T2T3 + TBAQ train & TBAQ test (20\%) & 0.87 & 0.87 & 0.87 & 80.1\% & 80.1\% & 80.1\% \\
\hline
\end{tabular}
\caption{Timex recognition results. TBAQ corresponds to the merger of the TimeBank 1.2 and AQUAINT TimeML corpora.}
\label{tab:results}
\end{table*}

\section{Resource creation}
\label{resource}

In this section we describe the results of converting all the aforementioned corpora. 
Table~\ref{tab:result-timeml} shows the volume of TIMEX3 and other TimeML element annotations created by conversion from TIMEX2. The TIMEX2 counts are generally slightly lower; this may be caused by the removal of nested temporal expressions, where the outer timex annotation is removed during conversion. To show the timex composition of the resultant corpora, Table~\ref{tab:result-types} shows the distribution of timex type in native and in converted corpora; they introduce 18~803 new TIMEX3 annotations.

Of these timexes, 4~343 are in ``web-grade" data -- that is, data taken from blogs, forums newsgroups and Wikipedia. These include 2~676 from Wikiwars (Wikipedia) and the remainder from ACE2005 -- 675 from newsgroups, 20 from community forums and 972 from crawled web text. This is a significant resource for developing automatic methods to accurately and consistently annotate temporal information for the semantic web.

\section{Evaluation}
\label{evaluation}

%- can we do some qualitative eval?

We evaluate the impact of our new resources by measuring the performance of a state-of-the-art timex recogniser, TIPSem-B~\cite{llorens2010tempeval-2}. It achieves competitive performance when trained over the TimeBank and AQUAINT corpora. We extend its training set to include our newly generated data. Our evaluation includes timex annotation (both recognition and interpretation) performance on:

\begin{enumerate}
\item New (T2T3) data when trained on prior data (TimeBank + AQUAINT) to show the ``difficulty" of the new data given current TIMEX3 training resources;
\item A mixture of prior and T2T3 data, with an 80/20 training/test split, to show how the recognition method handles the new data;
\item Prior data with an 80/20 training/test split, as a baseline measure;
\item As above, but with all of the T2T3 data added to the training set, to see its impact on the TIMEX3 task as previously posed;
%\item The mixed-genre TimenEval dataset~\cite{llorens2012timen}, with all prior annotation and all new T2T3 annotation used as training data, to compare with recognition performance on TimenEval using just prior annotation data for training.
\end{enumerate}

Performance is reported using both entity recognition precision and recall (strict), as well as the TempEval-2 scorer, which uses a token-based metric instead of entity-based matching (see \newcite{verhagen2010semeval} for details). Results are given in Table~\ref{tab:results}. 

\subsection{General recognition}

The timex recognition performance on our T2T3 data of systems trained using prior newswire-only corpora was low, with an F1 measure below 50\%. This suggests that existing resources are not sufficient to develop generic timex recognition models that are effective outside the newswire genre. However, existing recognition methods are capable of adapting to the new corpora given some of it as training data; an 80/20 training/test split of combined newswire/T2T3 timexes gave F1 measures in the seventies.

\subsection{Improving performance on newswire}

It is useful to measure performance on a TempEval-2-like task -- recognising timexes in the TimeBank/AQUAINT TimeML corpora. To this end, we set an 80/20 training/test split of TBAQ (TimeBank + AQUAINT) and measured system performance on a model learned from the training data. The large T2T3-generated resource is then added to the training data, the recognition model re-learned, and performance evaluated. As shown by the results, the larger set of more-diverse training data provides an improvement over the TimeBank set. Recall rises considerably at the cost of some precision, under both evaluation metrics. This matches with what one might expect given a much wider range of expression forms in the training data. The final result for TempEval-2 F1 measure is greater than the best score achieved during the TempEval-2 evaluation task.

\section{Conclusion}
\label{conclusion}

Identifying and overcoming issues with TIMEX2/TIMEX3 conversion, we have created a robust tool for converting TIMEX2 resources to TimeML/TIMEX3. Using this, we have generated a TIMEX3 resource with an order of magnitude more annotations than all previous resources put together. The resource contains new information about temporal expressions, and is helpful for training automatic timex annotation systems.

We have made both the transduction tool and the TIMEX3 annotated results available, as part of a public repository. Version 1.0 of the data is packaged as a single release available on the project web page (distribution licenses apply).

\subsection{Future Work}
As TIMEX2 resources exist in languages other than English (particularly Spanish and German), T2T3 can be enhanced to cater for these as well as English.
% The resources generated in this paper will help the temporal IE community.

Further, the extensive diversity of temporal expression phrasings found in the corpora introduced in this paper coupled with their TIMEX3 annotations is a significant boon to those working on the problem of timex normalisation.

\subsection{Acknowledgments}
The authors would like to thank Lisa Ferro of the MITRE Corporation for kind help with corralling TIMEX2 resources. The first author would also like to acknowledge the support of the UK Engineering and Physical Science Research Council in the form of a doctoral training grant. This paper has been also supported by the Spanish Government, in projects {\scriptsize TIN-2009-13391-C04-01}, {\scriptsize MESOLAP TIN2010-14860}, {\scriptsize PROMETEO/2009/119} and {\scriptsize ACOMP/2011/001}.

\bibliographystyle{lrec2006}
\bibliography{t2t3-corpus}

\end{document}